# Evaluation of Augmented Reality-based Intuitive Interface for Robot-Assisted Transesophageal Echocardiography: A User Study

Xiu Zhang*, Matteo Di Mauro*, Sofia Breschi, Angela Peloso, *Student Member, IEEE,* Emiliano Votta, Arianna Menciassi, *Fellow, IEEE*, and Elena De Momi, *Senior Member, IEEE*

***Abstract*—TransEsophageal Echocardiography (TEE) is essential for diagnosing and guiding Structural Heart Disease (SHD) interventions. However, manual TEE manipulation demands significant operator expertise, is physically demanding, and exposes clinicians to radiation when performed alongside fluoroscopy. Robotic-assisted TEE systems have been introduced to improve probe handling and reduce operator fatigue, yet the design of intuitive and effective user interfaces remains an open challenge. This study presents and evaluates a model-enhanced, Augmented Reality (AR)–based intuitive interface for robot-assisted TEE, designed to improve spatial awareness and control intuitiveness. A robotic TEE platform integrated with electromagnetic tracking and a virtual simulator was used to compare three user interfaces differing in visualization and interaction modalities: 2D joint-level (2D–JI), 3D joint-level (3D–JI), and 3D tip-level (3D–TI). Thirty-six participants performed standardized navigation tasks to reproduce target echocardiographic views, with performance assessed via position and orientation errors, completion time, and NASA–TLX workload scores. Results show that 3D visualization significantly improved spatial accuracy, reducing median position error from $13\ mm$ to $3\ mm$ and halving the orientation error compared with the 2D interface. Tip-level interaction yielded a further $50\%$ reduction in orientation error and reduced inter-user variability relative to joint-level control.Overall, the 3D–TI configuration, combining immersive visualization with direct tip-level control, proved the most effective and ergonomic interface, supporting the integration of AR-based visualization and intuitive control paradigms into next-generation robotic TEE systems to enhance operator performance and procedural safety.***

***Index Terms*—Augmented Reality, User Interface, Transesophageal Echocardiograph**

## I. Introduction

TRANSESOPHAGEAL echocardiography (TEE) plays a critical role in the diagnosis, planning and intraoperative monitoring of Structural Heart Diseases (SHDs) [1]. The American Society of Echocardiography advocates for the acquisition of a comprehensive TEE, which consists of the standard 28 echocardiographic views to better understand the underlying pathology and facilitate planning [2]. Performing TEE requires specialized training in both probe manipulation and image interpretation, and the steep learning curve makes examination quality highly dependent on the sonographer's experience. Moreover, its prolonged use can lead to patient discomfort and potential complications, such as esophageal trauma [3].

Real-time guidance during interventional SHD procedures, such as transcatheter mitral valve repair, left atrial appendage occlusion, and transcatheter aortic valve replacement, relies also heavily on TEE [4]–[6]. During these procedures, TEE is often performed alongside fluoroscopy, requiring sonographers to wear heavy lead aprons, which can lead to fatigue and increase the risk of long-term radiation exposure [7]. To overcome these challenges, recent advancements in teleoperated robotic-assisted TEE aim to enhance procedural ergonomics, reduce operator fatigue, and improve imaging consistency.

The TEE probe consists of a long gastroscopic tube with a transducer at the tip and a handle on the proximal end, which enables tip steering through a tendon-driven mechanism. Recent research has focused on developing add-on devices to automate the steering knots on the handle while also enabling translational and rotational movements [8]–[10]. Wang et al. proposed a teleoperation framework with virtual admittance that allows users to directly control through a gamepad [11]. However, precisely positioning a soft robotic probe over a long distance in a confined space remains challenging. To address this, Li et al. introduced a magnet-actuated TEE robot by attaching a permanent magnet to the probe tip, enabling direct tip control [12]. While these advancements improve maneuverability, there remains a gap in evaluating the efficiency and usability of the teleoperated interface for TEE, particularly in terms of user experience and clinical applicability.

Augmented Reality (AR) devices have been proven to be efficient tools in intervention procedures by providing 3D intuitive holograms [13]–[16]. For instance, the development of human-robot visual interfaces in neurosurgery, has shown that integrating intuitive joystick control with 3D visualization enables operators to follow complex trajectories with minimal training [17]. Moreover, Lin et al. developed ARei, an AR-assisted, touchless teleoperated robotic system for endoluminal procedures, which integrates gesture-based control, robot shape sensing, and preoperative anatomical models within an immersive AR scene [18]. In general, integrating AR 3D visualization can improve spatial awareness and reduce operator burden [19], [20].

Xiu Zhang is with Department of Electronics, Information and Bioengineering, Politecnico di Milano, Milan 20133, Italy, and also with the BioRobotics Institute, Scuola Superiore Sant'Anna, 56025 Pontedera, Italy (Corresponding author: Xiu Zhang) (xiu.zhang@santannapisa.it).

M. Di Mauro, A. Peloso, S. Breschi, E. Votta, E. De Momi are with the Department of Electronics, Information and Bioengineering, Politecnico di Milano, Milan 20133, Italy (matteo.dimauro, angela.peloso, sofia.breschi, emiliano.votta, elena.demomi)@polimi.it).

Arianna Menciassi is with the BioRobotics Institute, Scuola Superiore Sant'Anna, 56025 Pontedera, Italy (arianna.menciassi@santannapisa.it).

Xiu and Matteo contribute equally on this work.

This work was supported by the European Union's Horizon 2020 research and innovation program under the ARTERY grant agreement No. 101017140.

This work has been submitted to the IEEE for possible publication.

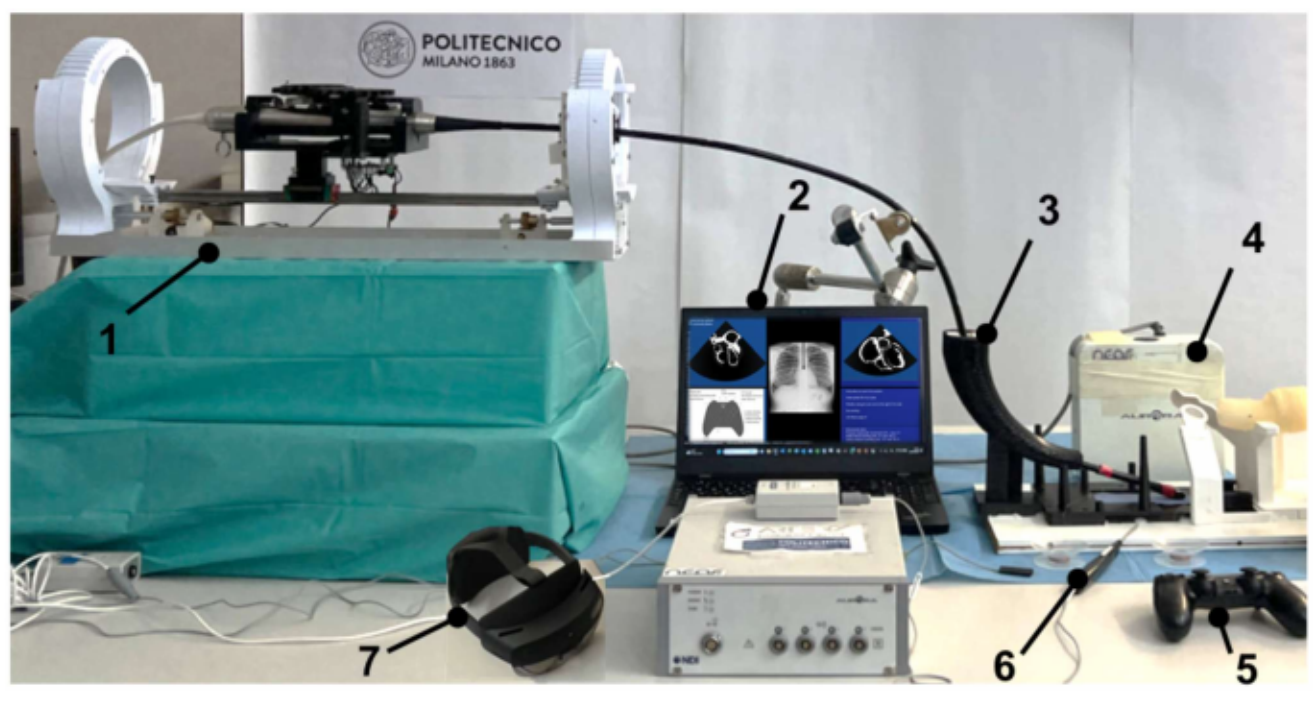

Fig. 1. Hardware setup for in-vitro user study: 1. Robotic TEE probe; 2. Standard 2D monitor; 3. Patient-specific esophagus phantom; 4.EM tracking system; 5. Gamepad; 6. EM probe; 7. Optical See-Through Head-Mounted Display (OST- HMD).

TEE works similarly to flexible endoscopy, where the tip movement is controlled by a combination of tube translation, rotation, and distal bending. This results in a non-intuitive mapping between the joint space and the actual motion of the probe tip [21], [22]. To overcome this challenge, Rozeboom et al. investigated intuitive user interfaces for flexible endoscope tip control, showing that robotic steering with touchpads and nonlinear joysticks improved navigation efficiency and reduced learning effort as compared to conventional methods [23]. Furthermore, Finocchiaro et al. proposed a comprehensive evaluation framework for human-machine interfaces (HMIs) in robot-assisted colonoscopy, highlighting how interface design directly impacts clinical performance and user stress [24]. However, unlike camera-based endoscopes that follow eye-in-hand paradigm, TEE provides visual feedback in the form of ultrasound images, which lack direct spatial correspondence to probe motion, making intuitive control even more challenging.

In this work, we propose and evaluate an AR-based intuitive interface for robot-assisted TEE procedures. The system comprises: (1) a robotic TEE platform integrated with an electromagnetic (EM) tracking system for real-time probe localization; (2) a virtual simulator that generates echocardiographic images; (3) three user interfaces designed to compare different visualization (2D vs. 3D) and interaction (joint-level vs. tip-level) modalities; (4) a user study incorporating both qualitative and quantitative assessments to evaluate usability, performance, and user experience. To the best of our knowledge, this study provides the first systematic evaluation of AR-enhanced interfaces for robotic TEE and offers insights into how visualization and control modalities impact operator performance during complex echocardiographic navigation tasks.

## II. Methodology

### A. System Components

*1) Hardware:* The experimental setup shown in Fig. 1 is composed of:

1) **Robotic TEE probe:** the robotic system comprises various components, including an TEE probe from Philips (X7-2t, Philips, Nederland) mounted on a robotic actuation system. The latter consists of two separate robotic systems: one controlling the flexion DoFs (ante-posterior and right-left), and the other controlling the translation in space and the rotation of the TEE probe along its longitudinal axes [25].
2) **Standard 2D monitor:** A laptop is used to display simulated fluoroscopic and echocardiographic images in a 2D format.
3) **Patient-specific esophagus phantom:** A physical phantom replicates the anatomical structures of the mouth with the bite-block, pharynx, and the esophagus, providing a realistic environment for navigation tasks.
4) **EM tracking system:** EM sensors (Aurora, NDI, Canada) are attached on the TEE probe to measure the probe pose in real-time.
5) **Gamepad:** the controller (PS2, SONY, Japan) is a wireless input device equipped with ten buttons to control the TEE probe.
6) **EM probe:** The EM probe is used to measure the position of reference pillars on the phantom to register the EM with the pre-operative anatomy model.
7) **Optical See-Through Head-Mounted Display (OST-HMD):** the device used is the (HoloLens2, Microsoft, United States). This device allows the visualization of holograms projected on the lenses and reflected in the users' eyes, creating an overlap of the virtual world with the real world. AR objects are placed in the environment, allowing users to interact with them using hand gestures, thanks to the hand tracking system.

*2) Middleware:* The Robot Operating System (ROS) is employed to establish communication among the hardware components. The EM tracking system acquires pose data at a frequency of 40 Hz and publishes it as messages on the ROS network. These messages are subsequently transmitted to the Unity environment via a TCP/IP communication protocol, implemented using the Unity Robotics Hub package. The robotic TEE probe is controlled at a frequency of 10 Hz.

### B. TEE Probe Kinematics

To find the optimal echo view, the operater positions a TEE probe by combining all 5-DoFs in the joint space, including: translating ($q_d$) and rotating ($q_\gamma$) along the axis direction by moving the TEE probe; rotation of the ultrasound plane ($q_\psi$); and bending in two directions, e.g. ante/retro ($q_{A-R}$) and left/right ($q_{L-R}$) directions through the tendon-driven mechanism controlled with knobs on the proximal handle (Fig. 2A).

The TEE robot works as a tendon-driven continuum robot, which can be modeled using a Constant Curvature (CC) model. As shown in Fig. 1B, frame 0 is defined as the origin of the robot. The base of the TEE probe is obtained by rotating along the z-axis by the roll angle ($\gamma$). Under the constant curvature assumption, the configuration of the bending section is defined by the rotation angle ($\theta$), the bending angle ($\alpha$), and the length of the bending section ($L$). The tip pose of the TEE probe is then derived by applying a rigid translation from the end-effector pose $[\boldsymbol{p}_e, \boldsymbol{x}_e, \boldsymbol{y}_e, \boldsymbol{z}_e]^T$ along the $z_e$ direction by a distance $L_{tip}$. Then, the rotation of the ultrasound scan plane along the $x_{tip}$ direction with the plane rotation angle ($\psi$) can define the scan direction. The forward kinematics is represented by the transformation matrix $\mathbf{T}(d, \gamma, \theta, \alpha, L, L_{tip}, \psi)$.

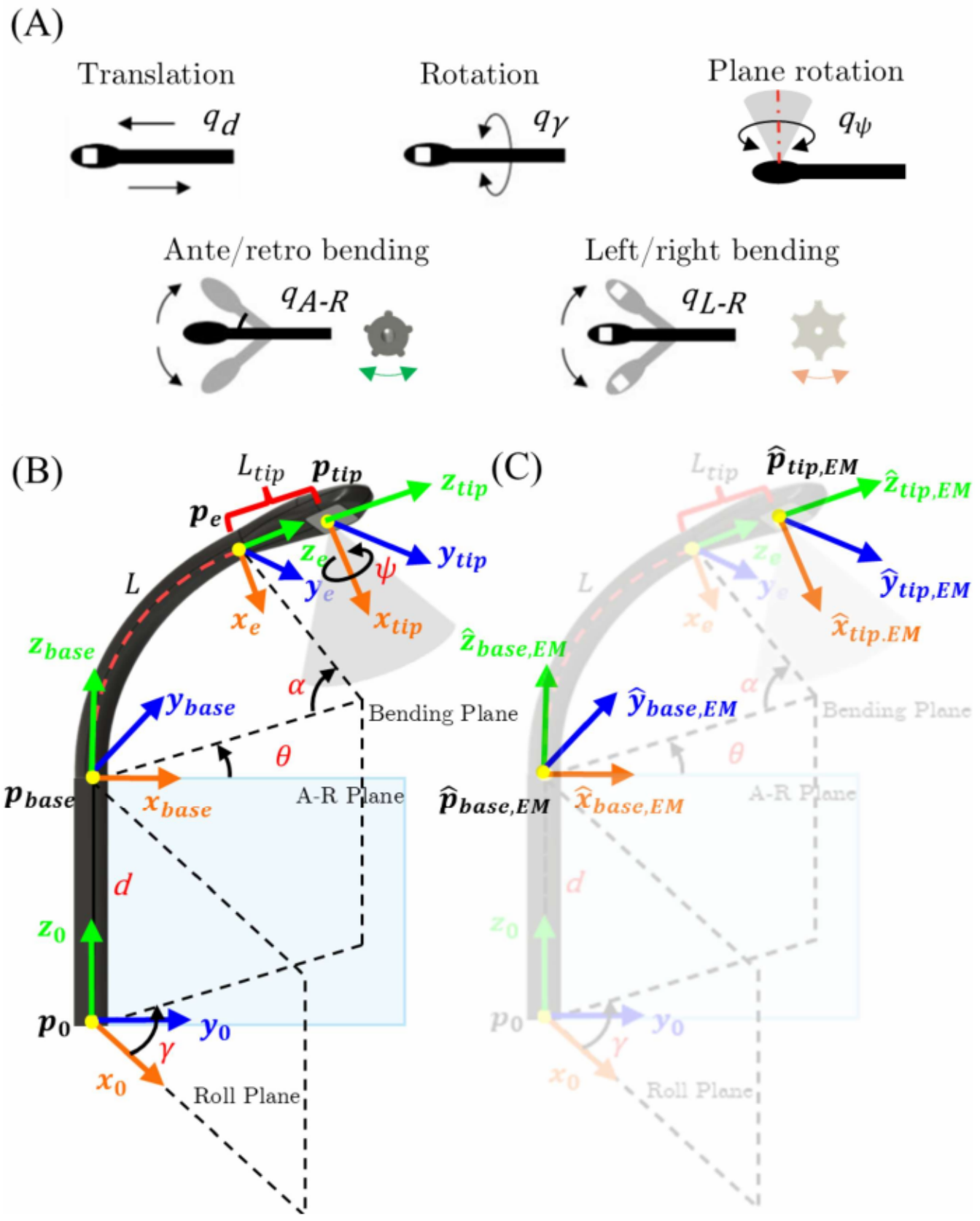

Fig. 2. Kinematics of the TEE probe: (A) Joint space degrees of freedom (DoFs). (B) Forward kinematics of the TEE probe include: translation ($d$), roll angle ($\gamma$), rotation angle between the A−R plane and the bending plane ($\theta$), bending angle ($\alpha$), length of the bending section ($L$), length of the tip ($L_{tip}$) and ultrasound plane rotation angle ($\psi$). (C) The pose measurements are acquired using EM sensors attached at both the tip and the base of the TEE probe.

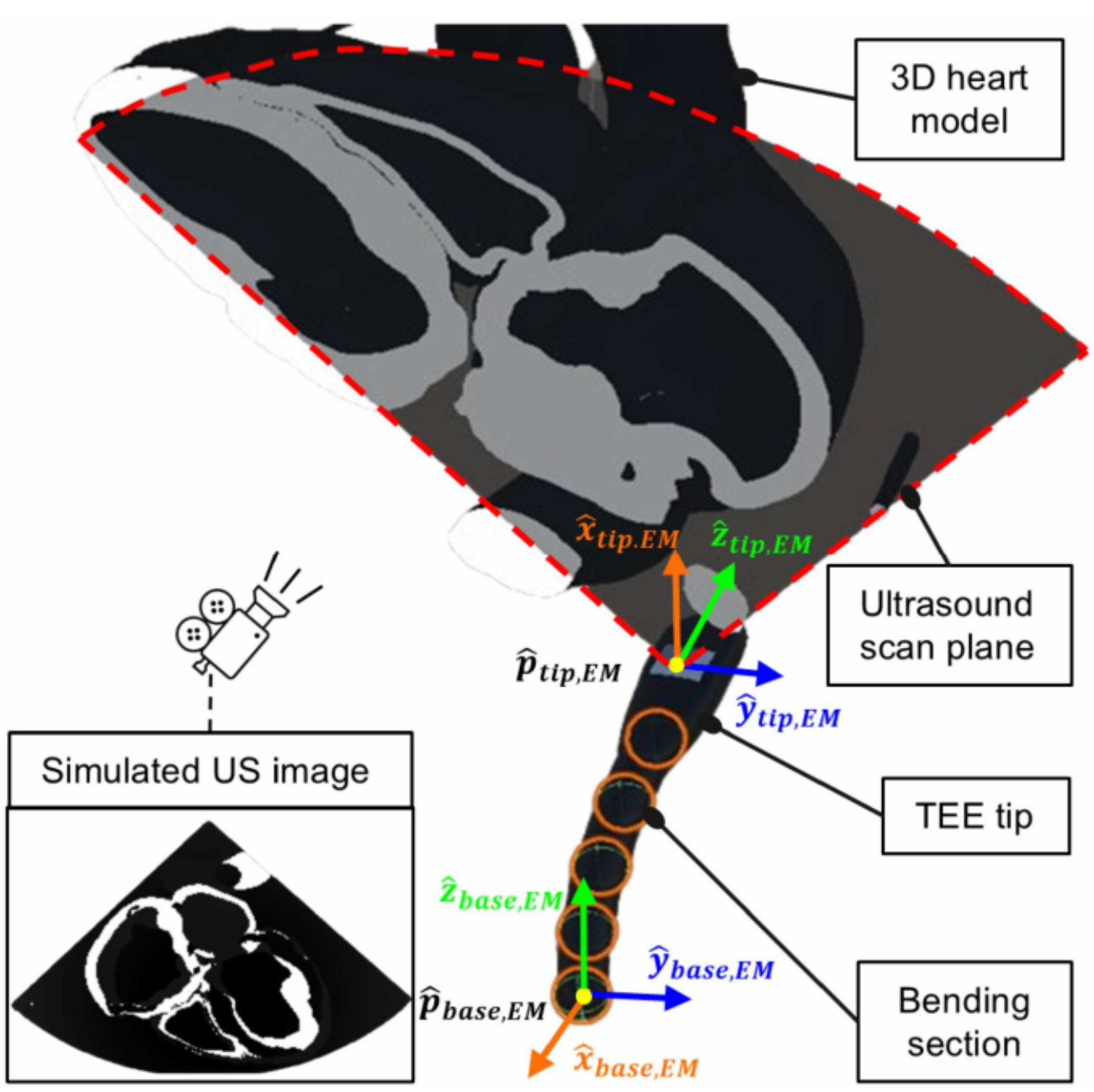

Fig. 3. Virtual reconstruction of the TEE probe within the simulator, along with the corresponding simulated US image.

Additionally, EM sensors, positioned at both the tip and the base of the TEE probe, measure poses in Cartesian coordinates, yielding $[\hat{\boldsymbol{p}}_{tip,EM}, \hat{\boldsymbol{x}}_{tip,EM}, \hat{\boldsymbol{y}}_{tip,EM}, \hat{\boldsymbol{z}}_{tip,EM}]^T$, $[\hat{\boldsymbol{p}}_{base,EM}, \hat{\boldsymbol{x}}_{base,EM}, \hat{\boldsymbol{y}}_{base,EM}, \hat{\boldsymbol{z}}_{base,EM}]^T \in \mathbb{R}^6$ (Fig. 1C). These measurements are subsequently converted to the configuration space to reconstruct the TEE 3D model through the inverse kinematics procedure following the method proposed by Degirmenci et al. [26].

### C. TEE Simulator

We developed a real-time digital twin of the TEE probe that reconstructs its motion in a virtual environment using the measured poses of both the tip and the base. The simulation framework consists of three main components: the bending section of the probe, a 3D model of the heart, and the simulated ultrasound (US) image (Fig. 3).

- **The bending section:** The bending section of the TEE probe is modeled in Unity (Unity Technologies, United States) using fifty spherical elements. This discretization provides a smooth visual approximation of the probe shaft while maintaining real-time performance. The probe kinematics follow the CC model, which allows the bending shape to be computed directly from the updated base and tip positions via forward kinematics.
- **The 3D model of the heart:** A Computed Tomography (CT) scan of a patient with mitral regurgitation (Ospedale San Raffaele, Milan, Italy) is used to reconstruct the heart chambers and associated vessels. The anatomical structures are manually segmented using 3D Slicer (Harvard University, United States). Subsequently, Fusion 360 (Autodesk, United States) is then used to refine the segmented surfaces and to generate the corresponding anatomical walls. This step is necessary because certain external boundaries, such as the epicardial surface and the walls of the aorta and veins, are not sharply defined in the CT images and require geometric reconstruction.
- **The simulated US image:** US imaging operates by transmitting high-frequency sound waves into tissue and detecting the echoes that return. In this way, the TEE probe produces a 2D slice cross-sectional view, corresponding to the plane of the ultrasound beam [27]. To replicate this imaging principle, a fan-shaped ultrasound scan plane with a 90° field of view was generated perpendicular to the probe tip, with 1-DOF, allowing rotation around the $\hat{\boldsymbol{x}}_{tip,EM}$ axis. A custom shader [28] is used to implement binary partitioning, dividing the 3D heart model into two regions. The region on one side of the plane was rendered invisible, while the other remained visible, thereby generating real-time cross-section views. A virtual camera, constrained to remain perpendicular to the scan plane, is used to display the simulated US image corresponding to the extracted cross-section of the 3D heart model.

### D. Actuation Plane Calibration

Ideally, the A−R and L−R bending should be performed in the sagittal and mediolateral planes. However, manufacturing imperfections and material fatigue can induce plastic torsion [29], resulting in a misalignment of the probe's actuation planes. In addition, EM sensors attached to the probe cannot be perfectly aligned with the true bending plane, further contributing to orientation errors. To address this issue, a calibration algorithm was developed to correct the misalignment (Algorithm 1).

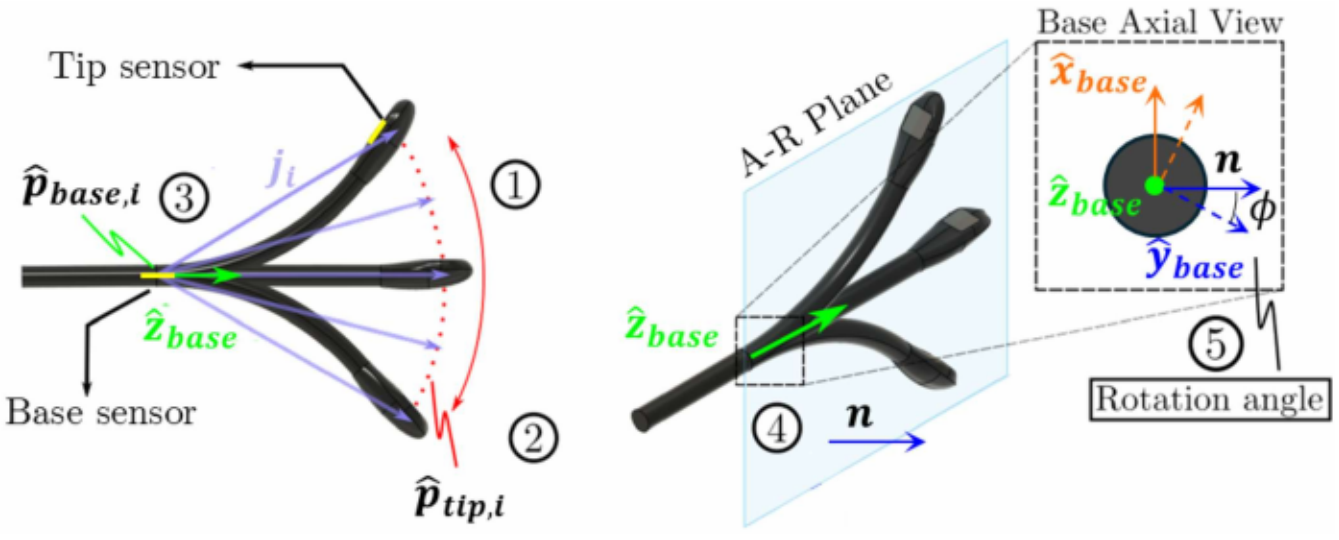

Fig. 4. Left: Lateral view of the probe with A-R movement, showing EM sensors at the tip and base. Right: Sensor orientation misalignment with bending plane. Numbers indicate the steps of the calibration algorithm.

On the TEE probe, two EM sensors are positioned at the tip and base of the bending section to measure the pose information in real time. As shown in Fig. 4, the calibration process begins by performing the A−R movement, while keeping the base fixed, During this process, the positions of the tip and base sensors for the $i$-th measurement are recorded as $\hat{\boldsymbol{p}}_{tip,i}(x_{tip,i}, y_{tip,i}, z_{tip,i})$ and $\hat{\boldsymbol{p}}_{base,i}$ $(x_{base,i}, y_{base,i}, z_{base,i})$ respectively. The relative displacement vector ($\boldsymbol{j}_i$) between these positions is calculated as:

$$\boldsymbol{j}_i = \hat{\boldsymbol{p}}_{tip,i} - \hat{\boldsymbol{p}}_{base,i} \tag{1}$$

These vectors are collected into a set $\mathbf{J} \in \mathbb{R}^{N\times3}$, which represents the relative motion of the tip sensor with respect to the base sensor during bending movements. Using the least squares method, the A−R actuation plane is identified from this dataset as the best-fit plane, where the normal vector $\boldsymbol{n}$ is determined as:

$$\boldsymbol{n} = \arg\min_{\boldsymbol{n}} \|\mathbf{J}\boldsymbol{n} + \boldsymbol{b}\|^2 \tag{2}$$

Here, $\boldsymbol{b}$ represents deviations of relative displacement vectors ($\boldsymbol{j}_i$) from the identified plane. Then, the normal vector, $\boldsymbol{n} \in \mathbb{R}^3$ of the measured A−R plane is computed using the Singular Value Decomposition (SVD) approach [30].

To align the probe's A−R actuation plane with the sagittal plane, a rotation matrix $\mathbf{R}_{z,base}(\phi)$ is defined. The angle $\phi$ is derived as following:

$$\phi = \arccos\left(\boldsymbol{n} \cdot \hat{\boldsymbol{y}}_{base}\right) \tag{3}$$

Here, $\boldsymbol{n}$ is the normal vector of the measured A−R plane, and $\hat{\boldsymbol{y}}_{base}$ is the unit vector in the y-direction as measured by the EM sensor at the base.

**Algorithm 1** Actuation Plane Calibration Algorithm

1: Perform continuous A−R bending.
2: Record data from a set of tip positions $\hat{\boldsymbol{p}}_{tip,i}$ and corresponding base positions $\hat{\boldsymbol{p}}_{base,i}$.
3: Compute the set of difference vectors $\mathbf{J} = \{\boldsymbol{j}_1, \boldsymbol{j}_2, ..., \boldsymbol{j}_N\}$
4: Identify the normal $\boldsymbol{n}$ that characterizes the plane that minimizes the distance to the set of vectors $\mathbf{J}$.
5: Derive the rotation matrix $\mathbf{R}_{z,base}(\phi)$ around the z-axis.

### E. User Interface Design

During the TEE examination procedure, the sonographer must teleoperate the robotic TEE probe to acquire specific US views. This study focuses on two main aspects: the development of the interaction modality and the visualization modality. Through the interaction devices, the user input is transmitted to the controller, which drives the robotic actuators. The corresponding probe motion is measured by the EM tracking system and mapped to a digital twin in the TEE simulator. Finally, the simulation output is delivered through visualization devices to provide visual feedback to the sonographer (Fig. 5).

*1) Interaction:* The joint-level interaction (JI) modality is the most commonly used in medical robotics, where interaction devices such as joysticks or buttons accept user input and convert it into binary signals (0 or 1). These signals are then mapped directly, in a one-to-one manner, to the joint space motions, including $q_d$, $q_\gamma$, $q_\psi$, $q_{A-R}$ and $q_{L-R}$.

In contrast, the tip-level interaction (TI) paradigm provides a more intuitive solution. In this approach, the user specifies the desired tip position, which is then converted into joint commands through the inverse kinematics of the TEE probe for robotic control in the joint space. In the proposed setup, the rotation motion ($q_\gamma$) is associated with the rotation angle of the probe base ($\gamma$) in the virtual environment. Moreover, the ultrasound plane rotation ($q_\psi$) is decoupled from the other motions and directly corresponds to the imaging plane angle ($\psi$). The 3D tip position vector ($p_{tip} \in \mathbb{R}^3$) can thus be mapped to the joint-space motions $[q_d, q_{A-R}, q_{L-R}]$ without introducing redundancy, ensuring a unique and well-defined inverse kinematics solution.

*2) Visualization:* Visualization plays a crucial role in robotic TEE, as the sonographer's ability to interpret anatomical structures and correlate them with probe motion directly affects procedural accuracy and efficiency. Two visualization modalities were implemented to evaluate their impact.

The first modality presents the probe motion and corresponding US views on a conventional 2D monitor. This approach resembles the current clinical workflow, where operators rely on planar images and mentally reconstruct the 3D anatomy by fusing information from different views. The second modality employs a 3D virtual hologram, in which the digital twin of the probe and the surrounding anatomical structures are rendered in real time.

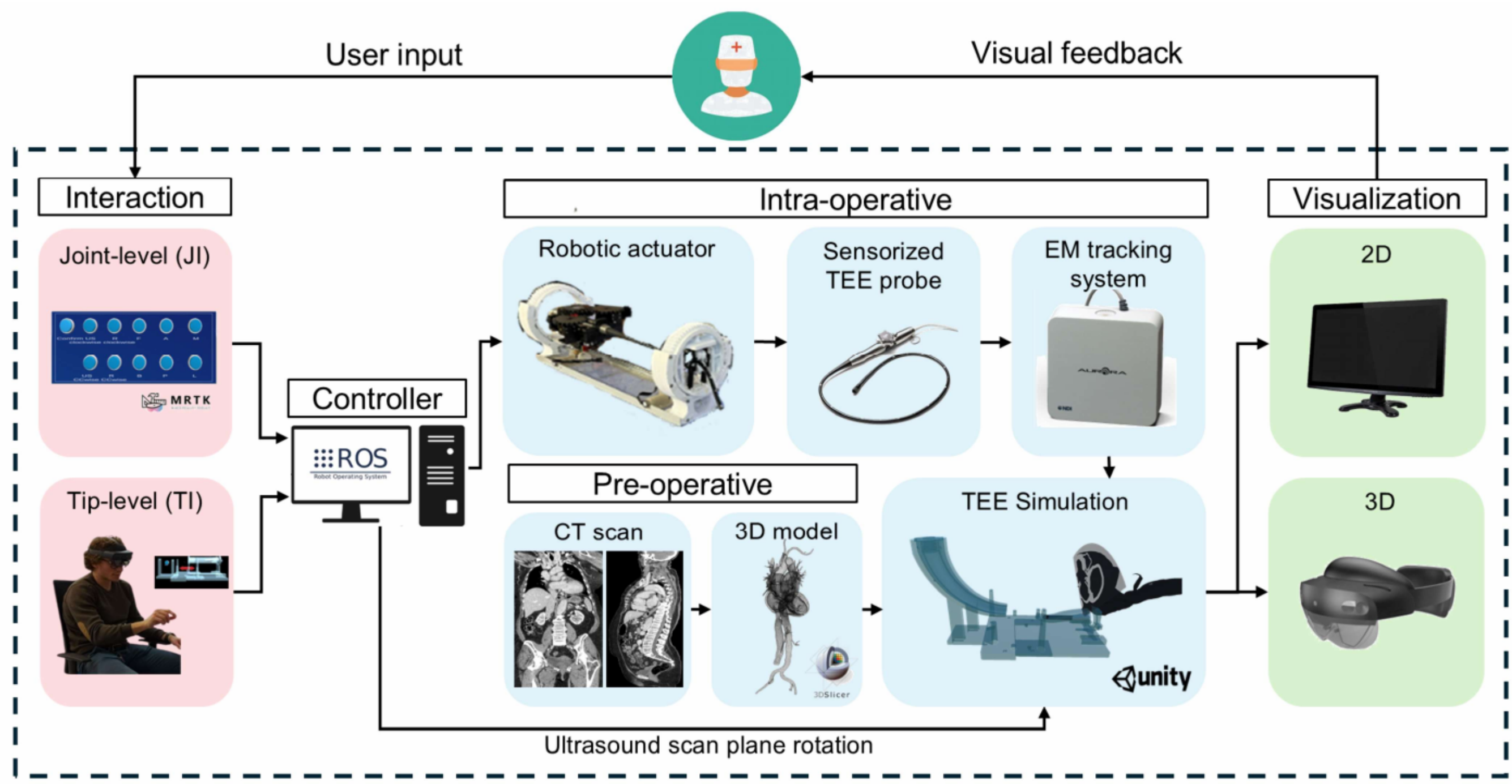

Fig. 5. User interfaces for robot-assisted transesophageal echocardiograph: User input is provided through joint-level or tip-level interaction devices and processed by the ROS-based controller. Intra-operatively, the robotic actuator drives a sensorized TEE probe, whose motion is tracked by an EM tracking system. Pre-operative CT data are reconstructed into 3D anatomical models and integrated into the TEE simulation environment. The resulting digital twin is visualized either on a conventional 2D monitor or within an immersive 3D holographic display, providing real-time feedback to the sonographer.

## III. Experiment

### A. Experimental setup

This study compared two interaction modalities (joint-level and tip-level) and two visualization modalities (2D and 3D). Participants were therefore divided into three groups: 2D–joint-level (2D-JI), 3D–joint-level (3D-JI), and 3D–tip-level (3D-TI). The comparison between 2D-JI and 3D-JI isolates the effect of visualization, while the comparison between 2D-JI and 3D-TI highlights the effect of interaction modality.

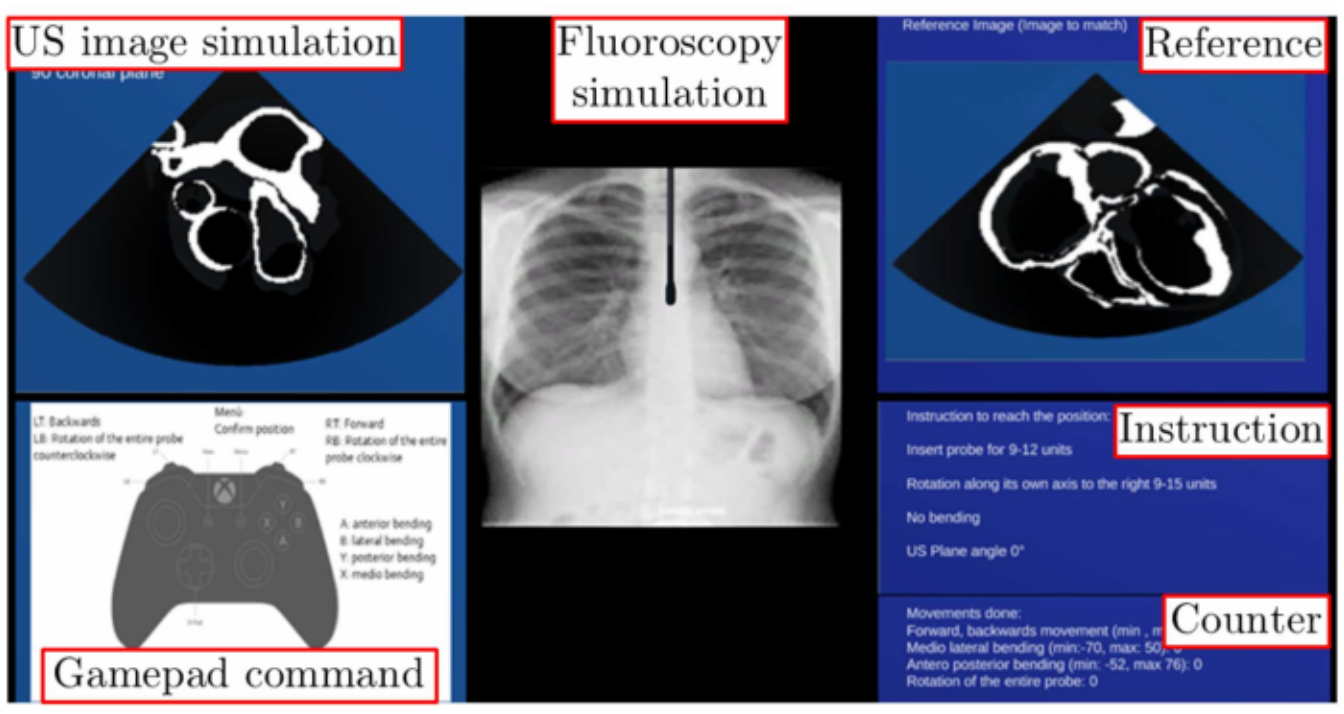

Fig. 6. Graphical user interface of the 2D-JI condition. The interface consists of five panels: (i) simulated ultrasound image (upper left), (ii) gamepad command panel (lower left), (iii) simulated fluoroscopic image with the virtual TEE probe (center), (iv) reference ultrasound image to be reproduced (upper right), and (v) instruction and counter panels providing task-specific guidance and real-time probe state information (lower right).

*1) **2D-JI**:* As in a conventional operating room setup, the sonographer received visual feedback from a 2D monitor connected to fluoroscopy. In the 2D-JI condition, participants similarly received feedback on a 2D monitor and controlled the joint-space motion of the TEE robot using a gamepad.

As shown in Fig. 6, fluoroscopy simulation was achieved by projecting the virtual bending section of the TEE probe onto a 2D fluoroscopic image. A simulated US image was displayed in the upper-left corner of the interface, while the panel below it showed the gamepad command instructions. On the right side, a reference image was provided as the target to be reached. The instruction panel offered task-specific guidance following standard TEE protocols. For example, to obtain a four-chamber view, the instructions included advancing the probe by $15 \sim 18$ units, rotating it by $5^\circ \sim 8^\circ$, bending it left by $30^\circ \sim 35^\circ$, and setting the ultrasound plane angle to $0^\circ$. Finally, a counter displayed the current joint states of the TEE robot.

*2) **3D-JI**:* In the 3D-JI condition, shown in Fig. 7, visualization was provided through an OST-HMD. This device projected holographic content in front of the user, enabling direct perception of the spatial pose of the simulated objects. Similar to the 2D-JI condition, four panels, i.e. the reference panel, ultrasound image simulation panel, instruction panel, and counter panel were integrated into the interface. In addition, an interactive command panel with eleven buttons replaced the physical gamepad, allowing participants to directly control the joint-space motions of the TEE robot.

Unlike the 2D monitor display, the OST-HMD enabled participants to physically move around while the 3D graphical interface remained stably positioned in their field of view. The TEE simulation was further enhanced by registering the virtual TEE probe to patient anatomy reconstructed from pre-operative CT data, thereby increasing anatomical realism. All interface panels were interactive and could be repositioned to a preferred location by simply dragging them within the augmented environment.

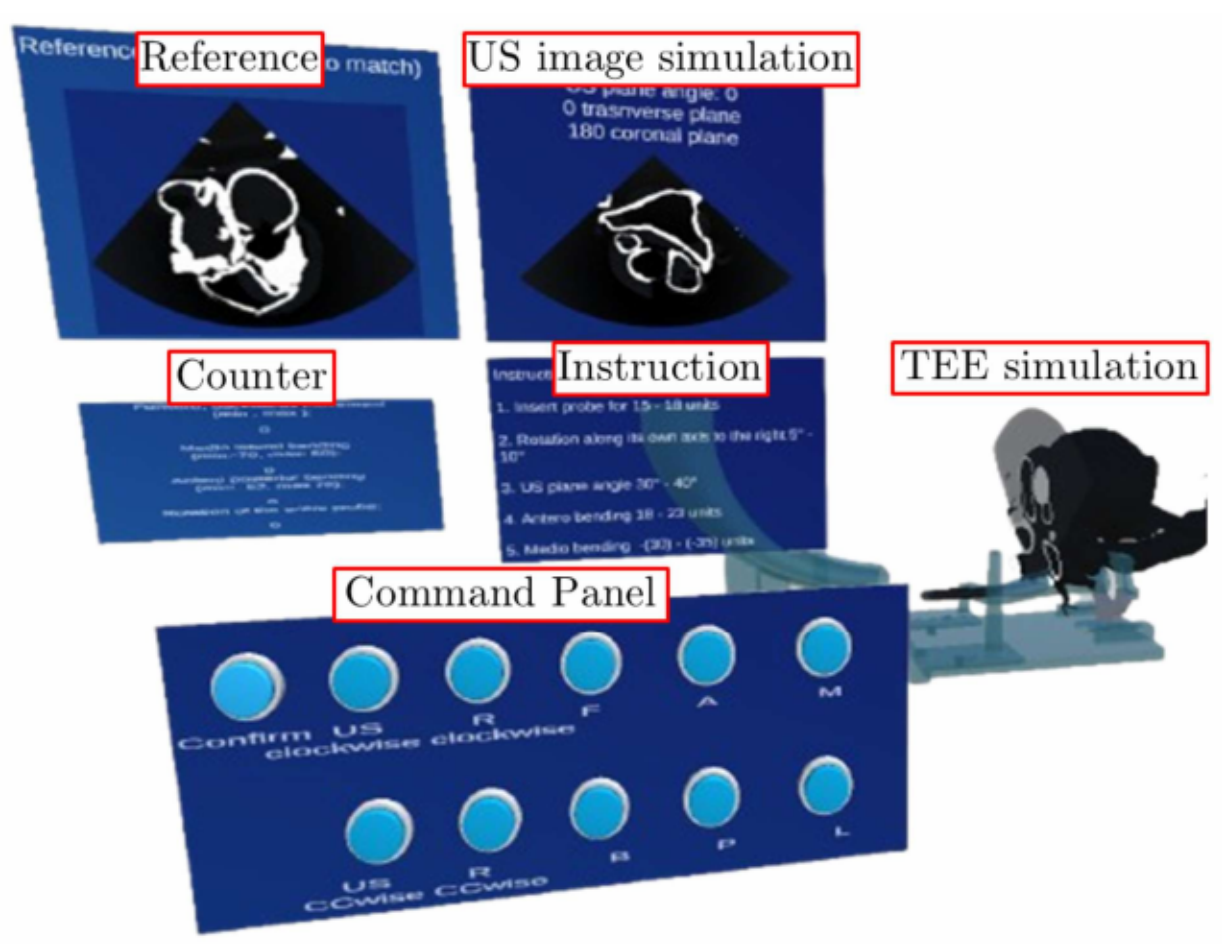

Fig. 7. Graphical user interface of the 3D-JI condition displayed through an OST-HMD. The interface includes the reference panel, US image simulation panel, instruction panel, counter panel, and an interactive command panel for controlling the joint-space motions of the TEE robot. The TEE simulation is registered onto the reconstructed patient anatomy.

*3) **3D-TI**:* In the 3D-TI condition, shown in Fig. 8, participants interacted directly with the virtual TEE probe within the holographic interface. Leveraging the inherent hand-tracking technology of the OST-HMD, participants could manipulate the system through natural gestures: the ultrasound scan plane could be rotated by grasping and turning its edge, while the TEE probe itself could be rotated by grasping its base. To adjust the probe's position, participants grasped the probe tip and moved it within the task space. This interaction paradigm provided a more intuitive and immersive method of maneuvering the probe compared with joint-level control.

As in the 3D-JI condition, the 3D-TI interface included four movable panels, e.g. the reference panel, ultrasound image simulation panel, instruction panel, and counter panel, that could be repositioned according to user preference. A confirmation button was also provided to record the final position of the probe tip once the desired view had been achieved.

### B. Experimental Procedure

A total of 36 participants were recruited in this study. They were between 20 and 35 years old and had mixed backgrounds in engineering, and they had no previous experience in the acquisition of TEE imaging. Participants were randomly assigned to three groups, with 12 individuals in each group.

At the beginning of the experiment, a short presentation was given, introducing the anatomical structure of the heart, the spatial relationship between the heart and the esophagus, and the principles of the TEE procedure. Following this introduction, each participant was given time to practice with their assigned user interface until they felt comfortable and confident operating it.

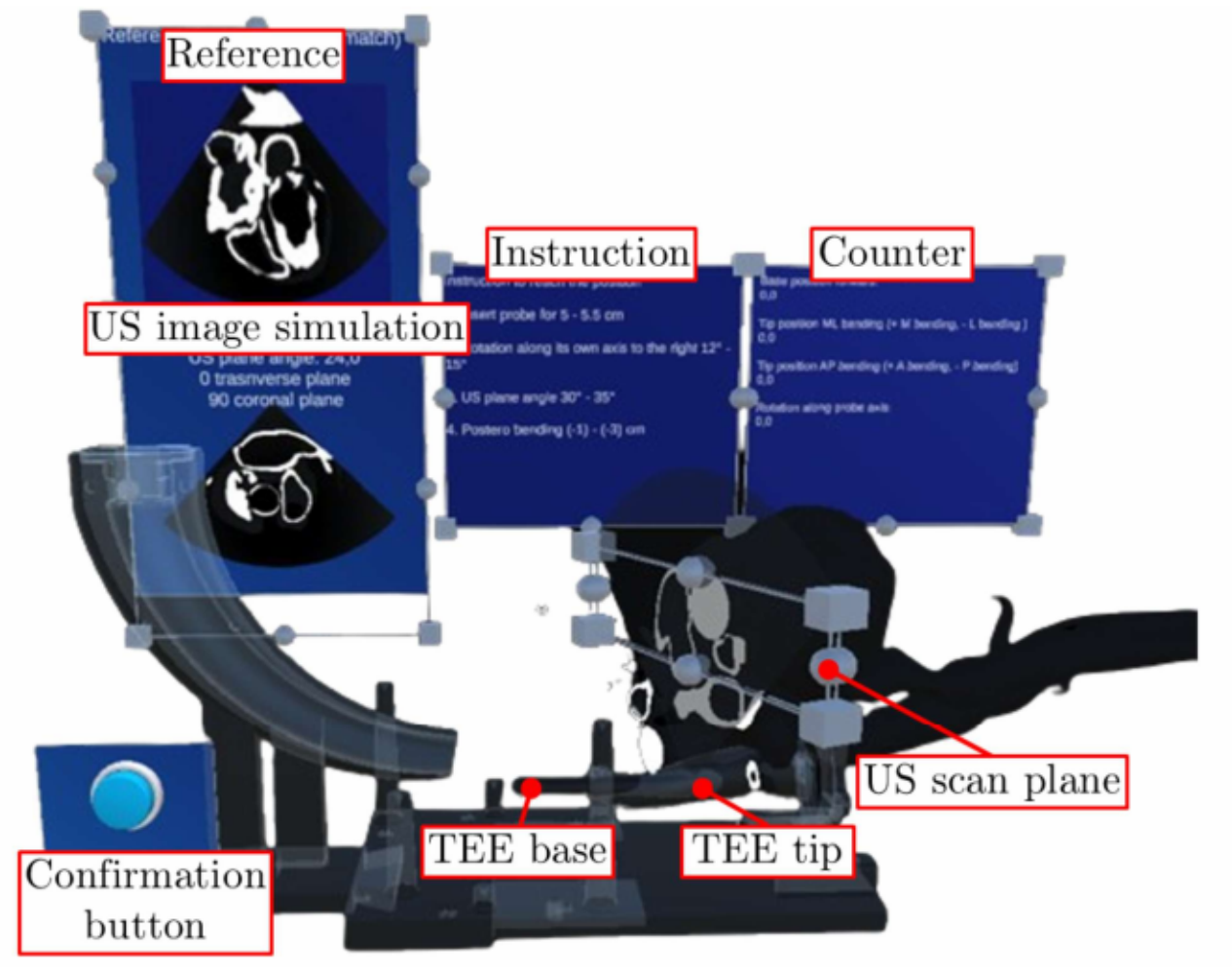

Fig. 8. Graphical user interface of the 3D-TI condition displayed through an OST-HMD. Participants could directly manipulate the virtual TEE probe by grasping the base, tip, or US scan plane using hand-tracking gestures. The interface also included four movable panels and a confirmation button used to record the final probe tip position once the desired view was achieved.

The experimental task consisted of controlling the robotic TEE system using the assigned interface to reproduce a reference US image. Three reference images were selected from the 28 standard TEE views, including two four-chamber views and one two-chamber view acquired from the mid-esophageal position (Fig. 9). Their order was randomized to minimize potential learning effects. For each reference image, participants performed two trials, resulting in six trials per participant. During each trial, participants manipulated the probe until the simulated US image matched the reference image as closely as possible. Once satisfied with the alignment, they confirmed the probe's position to complete the trial.

At the end of each trial, both the pose of the TEE probe and the trial completion time were recorded for subsequent performance analysis. After all six trials, participants completed the NASA Task Load Index (NASA-TLX) questionnaire [31] to evaluate perceived workload. The NASA-TLX consists of six subscales scored on a 21-point scale, ranging from 0 (very low workload) to 20 (very high workload).

### C. Performance Metrics

*1) Quantitative Evaluation:* For each reference image, the desired US plane pose was defined by combining the TEE tip position with the ultrasound plane rotation angle $\varphi_d$. This resulted in the desired tip pose $\hat{\boldsymbol{y}}'_{tip,d}$. After each trial performed by a participant, the achieved US plane pose was recorded as $\hat{\boldsymbol{y}}'_{tip,r}$. By comparing these two poses (Fig. 10), the performance could be objectively evaluated using the position error $E_p$ and the orientation error $E_o$, defined as follows:

$$E_p = \|\hat{\boldsymbol{p}}_{tip,d} - \hat{\boldsymbol{p}}_{tip,r}\| \tag{4}$$

$$E_o = \arccos\left(\frac{\hat{\boldsymbol{y}}'_{tip,d} \cdot \hat{\boldsymbol{y}}'_{tip,r}}{\|\hat{\boldsymbol{y}}'_{tip,d}\| \cdot \|\hat{\boldsymbol{y}}'_{tip,r}\|}\right) \tag{5}$$

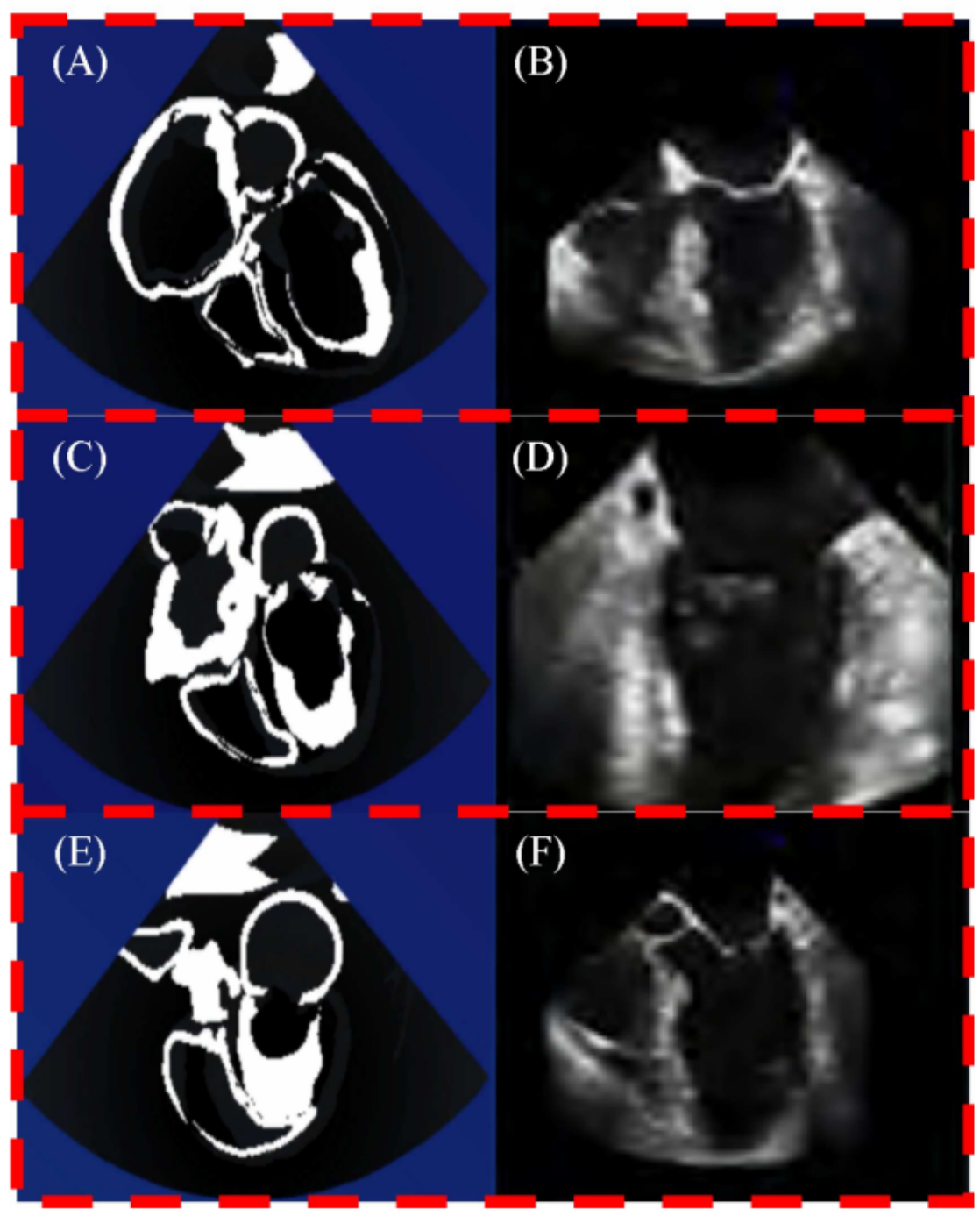

Fig. 9. Experimental task. Left column: Reference images (A, C, E). Right column: corresponding real clinical US images (B, D, F). Participants were instructed to manipulate the TEE probe until the simulated US image matched the given reference image.

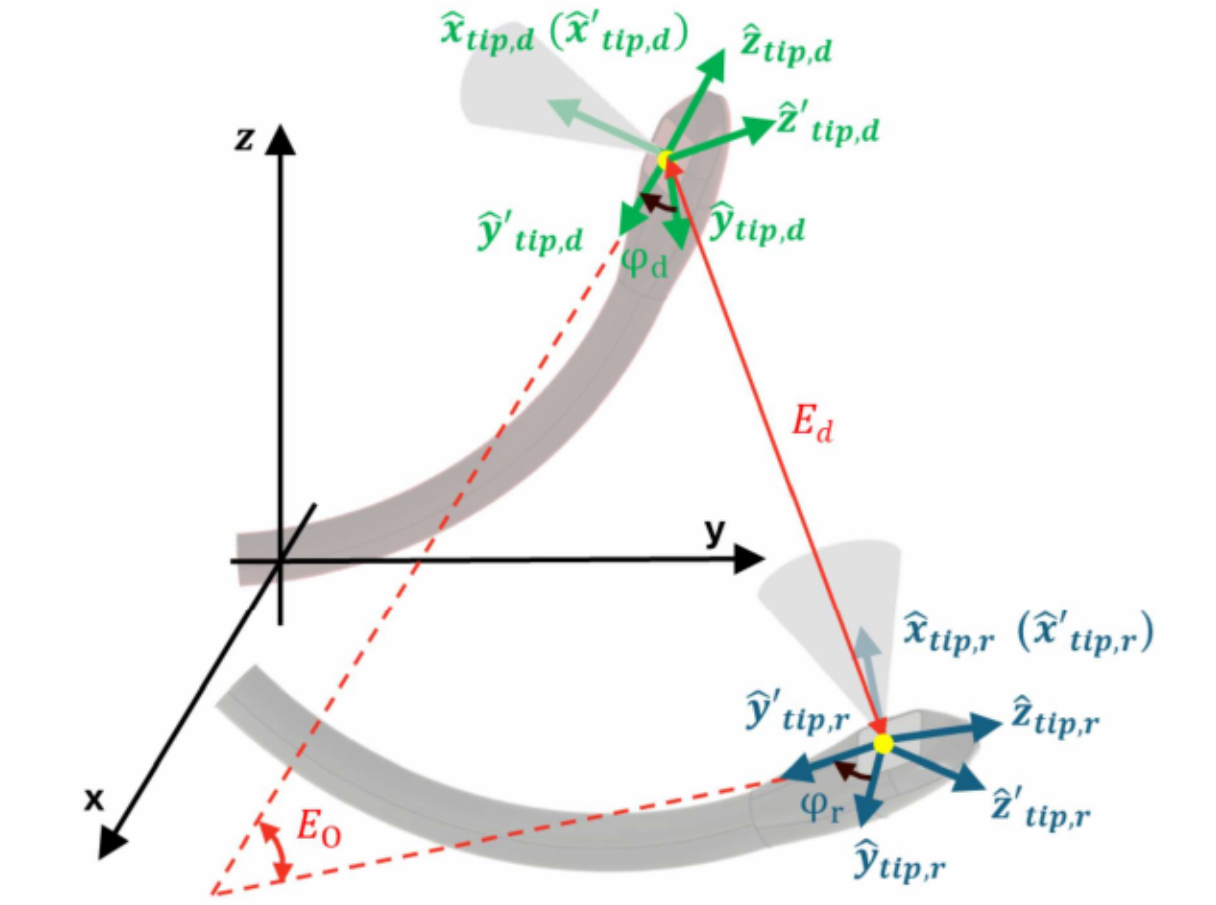

Fig. 10. Illustration of performance metrics used in the evaluation. The desired US plane pose $(\hat{x}_{tip,d}, \hat{y}_{tip,d}, \hat{z}_{tip,d})$ with rotation angle $\varphi_d$ (green) is compared with the achieved pose $(\hat{x}_{tip,r}, \hat{y}_{tip,r}, \hat{z}_{tip,r})$ with rotation angle $\varphi_r$ (blue). The position error $E_p$ is defined as the Euclidean distance between the desired and achieved TEE probe tip positions, while the orientation error $E_o$ corresponds to the angular difference between the two US imaging planes.

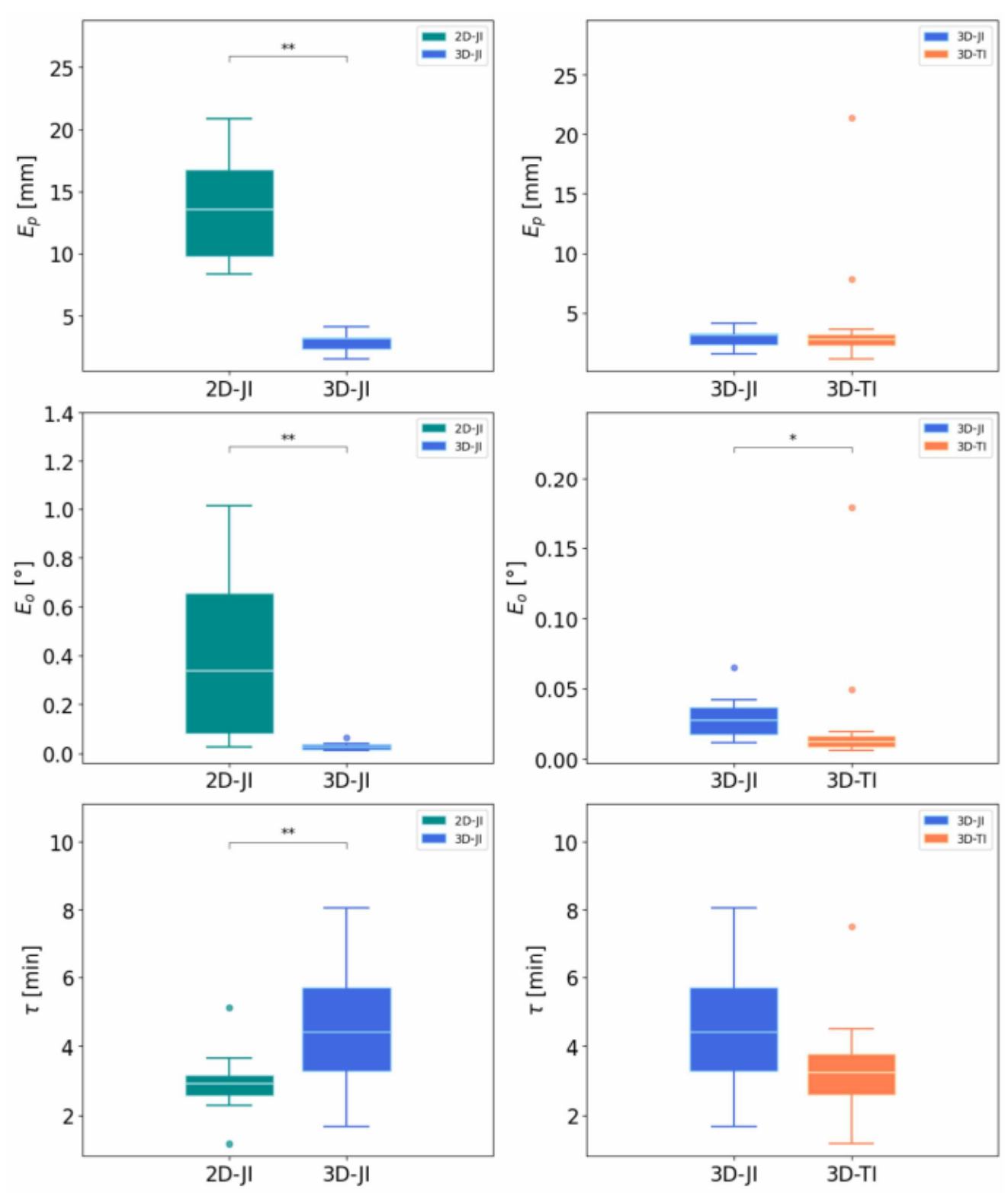

Fig. 11. Quantitative comparison of user performance across different visualization and interaction modalities. The left column compares 2D and 3D visualization (2D-JI vs. 3D-JI), while the right column compares interaction modalities (3D-JI vs. 3D-TI). ($*$, $p-value < 0.05$; $**$, $p-value < 0.01$)

In addition, a time counter recorded the trial duration $\tau$, which was used to evaluate how quickly participants were able to reproduce the reference image. The Shapiro–Wilk test was applied to each group (2D-JI, 3D-JI, and 3D-TI) and each performance metric (position error $E_p$, orientation error $E_o$, and duration time $\tau$) to examine data distributions.

*2) Qualitative Evaluation:* In addition to objective metrics, subjective workload was assessed for each participant using the NASA-TLX questionnaire. This questionnaire evaluates perceived workload across six subscales:

1) Mental Demand (MD). How mentally demanding was the task?
2) Physical Demand (PD). How physically demanding was the task?
3) Temporal Demand (TD). How hurried or rushed was the pace of the task?
4) Performance (PE). How successfully were you in accomplishing what you were asked to do?
5) Effort (EF). How hard did you have to work to accomplish your level of performance?
6) Frustration (FR). How insecure, discouraged, irritated, stressed and annoyed were you?

## IV. Results and Discussion

### A. Quantitative Evaluation

All metrics across groups were found to be non-normally distributed. Accordingly, Mann–Whitney U tests were used for pairwise comparisons between the 2D-JI and 3D-JI groups, as well as between the 3D-JI and 3D-TI groups.

*1) Effect of Visualization Modality (2D-JI vs. 3D-JI):* Figure 11 (left column) illustrates the quantitative comparison between 2D and 3D visualization interfaces. Across all metrics, 3D visualization markedly improved task accuracy. As compared to the 2D-JI modality, in the 3D-JI one the median value of $E_p$, was significantly reduced from a median of 13 $mm$ to 3 $mm$, and the median of the orientation error, $E_o$, decreased by 50%. However, the time-expense of probe adjustment increased from 3.2 $min$ to 4.5 $min$. Participants operating in the 3D environment required significantly more time to reproduce the target views than those using the conventional 2D interface. This increase in time expense is likely attributable to the participants' adaptation to the AR environment and the relative unfamiliarity of gesture-based operation compared to traditional joystick control.

*2) Effect of Interaction Modality (3D-JI vs. 3D-TI):* The right column of Fig. 11 compares the two interaction modalities. No statistically significant difference was observed in the position error, $E_p$, between the 3D-JI and 3D-TI conditions. Nonetheless, the orientation error, $E_o$, exhibited a significant reduction by 50% in the 3D-TI modality compared with 3D-JI, indicating superior control over the probe orientation when direct tip manipulation was available. Although the completion time, $\tau$, did not differ significantly between the two modalities, the interquartile range (IQR) for 3D-JI (2 $min$ to 8 $min$), was notably broader than that of 3D-TI (1 $min$ to 4 $min$). This suggests higher variability among participants using joint-level control, whereas the tip-level interface provided a more intuitive and consistent control experience across users.

### B. Qualitative Evaluation

Overall, participants reported comparable mental and temporal demands, perceived performance, effort, and frustration across all interface configurations. The only statistically significant difference was observed in the Physical Demand subscale, where the 3D–JI condition yielded higher scores than the 2D–JI. This suggests that the combination of the HMD and manual command input in the 3D–JI interface required greater physical effort, likely due to the need to maintain head posture and perform repeated button selections in the AR environment.

Despite the slightly increased physical demand, participants did not report elevated frustration or reduced perceived performance under the 3D interface. This suggests that users perceived the immersive visualization as beneficial rather than burdensome, even though it required additional physical effort. The alignment between qualitative workload ratings and the quantitative findings is also noteworthy: while the 3D visualization improved accuracy, qualitative ratings indicate that this improvement did not come at the expense of increased mental load. This supports the conclusion that the AR environment enhanced spatial understanding in an intuitive manner.

## V. Conclusion

This study evaluated a model-enhanced, AR–based intuitive interface for robot-assisted TEE through a user study comparing different visualization and interaction modalities. Quantitative and qualitative analyses strongly suggested that 3D visualization significantly improved spatial accuracy compared with the conventional 2D interface, reducing both position and orientation errors. The comparison between interaction modalities revealed that tip-level interaction achieved comparable accuracy as compared to joint-level control while enabling faster and more consistent performance across participants. In summary, the 3D–TI configuration, combining immersive visualization with intuitive and direct manipulation, proved to be the most effective interface for robotic TEE navigation.

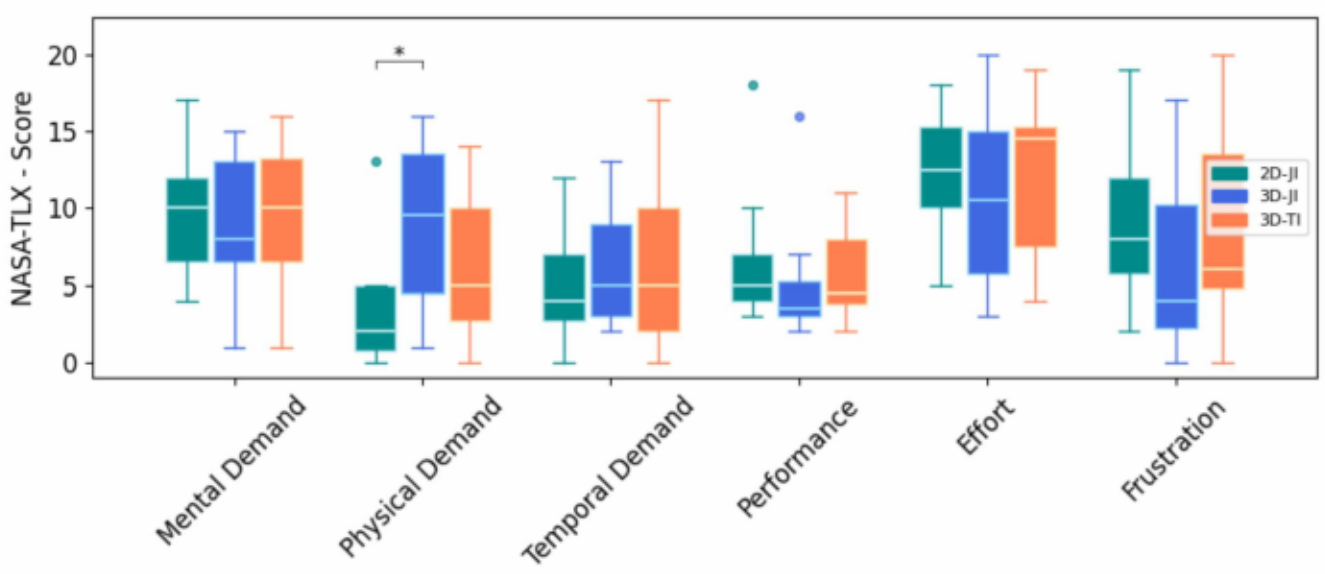

Fig. 12. Results of the NASA-TLX questionnaire assessing subjective workload across different visualization and interaction modalities. Each box represents the distribution of scores for the six workload dimensions: Mental Demand, Physical Demand, Temporal Demand, Performance, Effort, and Frustration. $*$, $p-value < 0.05$)

While these findings are encouraging, the experiments were conducted in a simulated environment using a phantom-based digital twin, which may not fully reflect the difficulty of interpreting echo imaging of clinical TEE procedures. Future work will focus on extending this framework in clinical settings. Specifically, we plan to integrate real echo imaging for live feedback and image-based navigation, and to develop registration algorithms that align the digital twin with patient-specific anatomy, enabling seamless interaction between the virtual and physical environments.

## VI. Acknowledgment

The experimental protocol for this user study was reviewed and approved by the Ethics Committee of Politecnico di Milano, Italy (Approval No. 45/2023). All participants were informed of the study objectives and procedures and provided written informed consent prior to participation.

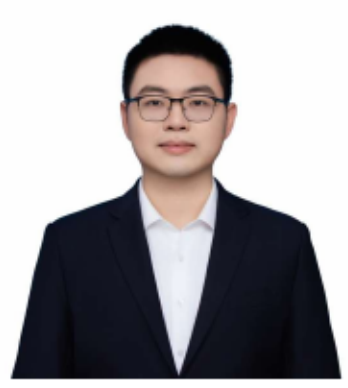

**Xiu Zhang** Xiu Zhang received the M.Sc. degree in Mechanical Engineering from Politecnico di Torino and the Ph.D. degree in Bioengineering from Politecnico di Milano. He is currently a postdoctoral researcher in the Surgical Robotics and Allied Technologies Area at the BioRobotics Institute, Scuola Superiore Sant'Anna, Italy. His research focuses on robotic systems for image-guided cardiovascular interventions, with particular interests in autonomous control, reinforcement learning-based planning.

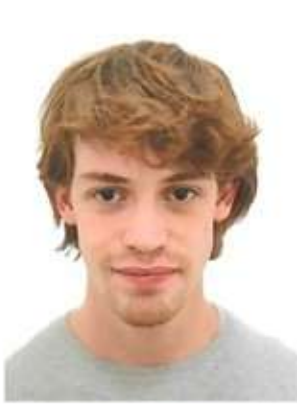

**Matteo Di Mauro** Matteo Di Mauro received the M.S. degree in Biomedical Engineering from Politecnico di Milano in 2024. He is currently pursuing the Ph.D in Artificial Intelligence in Biomedical Engineering, focusing on early detection and prevention of cardiac arrest using neural networks and deep learning techniques.

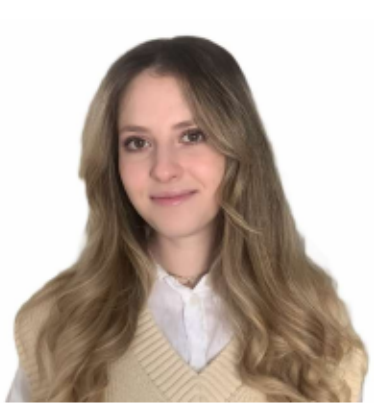

**Sofia Breschi** Sofia Breschi received her MSc in Biomedical Engineering from Politecnico di Milano, where she is currently pursuing her PhD. Her research focuses on mixed reality applications for robotic surgery, with a particular interest in surgical navigation and human–machine interaction.

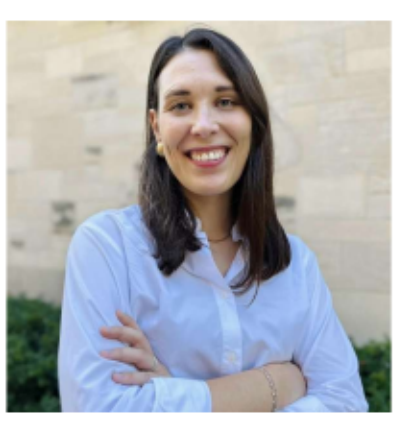

**Angela Peloso** Angela Peloso obtained MSc in Biomedical Engineering in 2021 at University of Padova. From 2021 to 2022 she was a research fellow at the University of Verona. She is currently enrolled as PhD Candidate at Politecnico di Milano. Her research has focused on path planning for autonomous steerable catheters under ARTERY project, and motion planning for continuum robotics.

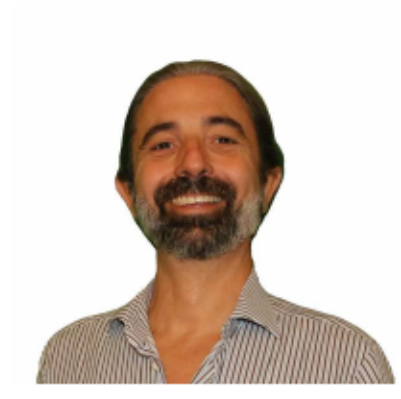

**Emiliano Votta** Emiliano Votta received his PhD in bioengineering in 2006 from Politecnico di Milano, where he is currently an associate professor at the Department of Electronics, Information, and Bioengineering. He is an expert in cardiovascular biomechanics and computer modeling, with an emphasis on numerical simulations, mixed reality, and image analysis. He is among the co-founders of Artiness srl, a spin-off company stemmed from Politecnico di Milano and devoted to the development of mixed reality solutions for interventional cardiology. He is also one of the co-inventors of a filed Italian patent.

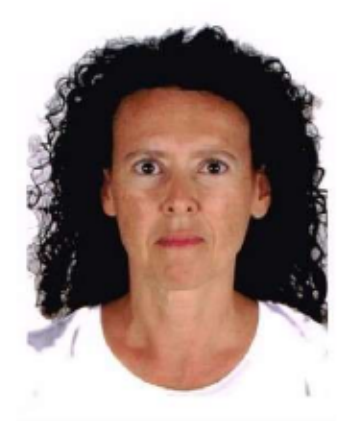

**Arianna Menciassi** Arianna Menciassi is a Full Professor of Biomedical Engineering at Scuola Superiore Sant'Anna (Pisa, Italy). She leads the Surgical Robotics and Allied Technologies Area. Her research focuses on the study and development of surgical and therapeutic robotic tools, procedures guided by ultrasound and magnetic fields, microrobotic interventions, and robotic artificial organs.

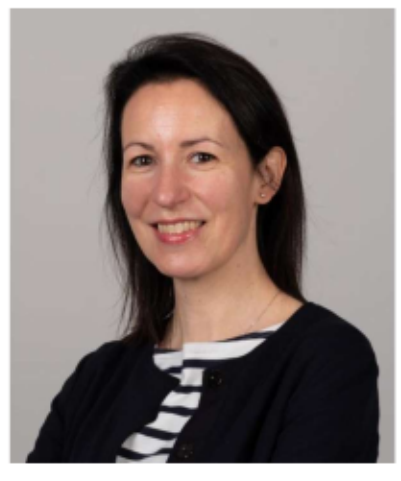

**Elena De Momi** Elena received her MSc in Biomedical Engineering in 2002, PhD in Bioengineering in 2006, and she is currently Full Professor in the Electronic Information and Bioengineering Department (DEIB) of Politecnico di Milano. She is co-founder of the Neuroengineering and Medical Robotics Laboratory, in 2008, being responsible of the Medical Robotics section.